\documentclass[letterpaper, 10 pt, conference]{ieeeconf}  
\usepackage{etoolbox}
\makeatletter
\patchcmd{\@makecaption}
  {\scshape}
\IEEEoverridecommandlockouts                              

\usepackage{microtype}
\usepackage{array}
\usepackage{graphicx}
\usepackage{subfigure}
\usepackage{booktabs} 
\usepackage{algorithmic}
\usepackage{algorithm}
\usepackage{amsmath}
\usepackage[backend=bibtex]{biblatex}
\usepackage{hyperref} 
\usepackage{makecell}
\usepackage{multirow}

\hypersetup{
     colorlinks   = true,
     citecolor    = green
}

\DeclareMathOperator*{\argmax}{arg\,max}

\newcommand\sbullet[1][.5]{\mathbin{\vcenter{\hbox{\scalebox{#1}{$\bullet$}}}}}

\graphicspath{{plots/}}
\let\labelindent\relax
\usepackage{enumitem}
\setlist[enumerate]{wide=0pt, leftmargin=15pt, labelwidth=15pt, align=left}
\title{\LARGE \bf
Meta Adaptation using Importance Weighted Demonstrations
}

\author{Kiran Lekkala$^{1}$, Sami Abu-El-Haija$^{2}$ and Laurent Itti$^{1}$
\thanks{This work was supported by the National Science Foundation (grants CCF-1317433 and CNS-1545089) and Intel Corporation. The authors affirm that the views expressed herein are solely their own, and do not represent the views of the United States government or any agency thereof.}
\thanks{Correspondence to Kiran Lekkala, {\tt\small klekkala@usc.edu}}
\thanks{$^{1}$Kiran Lekkala and Laurent Itti are with ILab, Department of Computer Science,
        University of Southern California, Watt way, Los Angeles, CA}%
\thanks{$^{2}$Sami Abu-El-Haija is with the ISI, University of Southern California,
        Los Angeles, CA 90007, USA}
}

\bibliography{main}
\begin{document}
\maketitle

\begin{abstract}
Imitation learning has gained immense popularity because of its high sample-efficiency. However, in real-world scenarios, where the trajectory distribution of most of the tasks dynamically shifts, model fitting on continuously aggregated data alone would be futile. In some cases, the distribution shifts, so much, that it is difficult for an agent to infer the new task. We propose a novel algorithm to generalize on any related task by leveraging prior knowledge on a set of specific tasks, which involves assigning importance weights to each past demonstration. We show experiments where the robot is trained from a diversity of environmental tasks and is also able to adapt to an unseen environment, using few-shot learning.  We also developed a prototype robot system to test our approach on the task of visual navigation, and experimental results obtained were able to confirm these suppositions.
\end{abstract}

\section{Introduction}
\label{sec:introduction}
In recent years, we have seen many agents perform numerous tasks using Imitation learning, in countless applications, especially robotics. There has been a significant progress made for algorithms which learn amidst noisy environments \cite{DBLP:conf/uai/FoxPT16}, sparse training signals \cite{pathakICMl17curiosity}, and imperfect demonstrations \cite{DBLP:conf/iclr/GaoXLYLD18}. However, there has not been much focus on allowing these agents to gather data and generalize to a wide variety of environments.

Especially for the task of navigation, this is quite crucial because, autonomous navigation systems like self-driving cars, delivery robots should be able to function in almost any situations \cite{\cite{DBLP:conf/icra/LekkalaI21, lekkala2016accurate, lekkala2015artificial, lekkala2014pid, lekkala2016simultaneous}
}. Since the data distribution continuously changes, it is challenging to learn a task from a fixed set of data, nor is it practical to obtain a comprehensive dataset \cite{DBLP:conf/icra/CodevillaMLKD18}. In nearly every real-world applications, the data distribution is long-tailed, meaning that the agent would always encounter new patterns, which has a small number of examples. There would always be instances where the agent has never encountered in the past. Taking a step forward, we would then want these systems to perform well in any given situation by applying prior patterns. Although we are concerned about navigation, researchers from other backgrounds can also find related reasons to concur with us.

Many of the existing solutions restrict their data domain, by training and testing on datasets collected on same environments \cite{\cite{wen2022can, xu2022ferroelectric, DBLP:journals/corr/abs-2305-15591, lekkala2020attentive}}. Other works like \cite{DBLP:journals/corr/BojarskiTDFFGJM16}, apply their algorithm to self-driving cars, have a much broader data domain. However, since these models were not trained in different settings, for example, cluttered, pedestrian-rich environment, they would not generalize to other settings. Some of the recent works, which try to generalize to new contexts are quite promising but also have some loopholes. With all these practical considerations, it is imperative that we design a method which enables the algorithm to function in diverse scenarios.

Meta-Learning deals with applying prior knowledge from various skills to learn a new skill in a few shot setting. These algorithms facilitate the model to utilize previous experience by constructing reusable structured patterns which could then be adapted in new contexts. We propose a method which meta-learns a set of tasks and generalizes to new tasks using a few samples. This paper deals with the first step in making an agent adapt to dynamically changing environments. 

\begin{center}
\begin{figure}
   \includegraphics[width=\columnwidth]{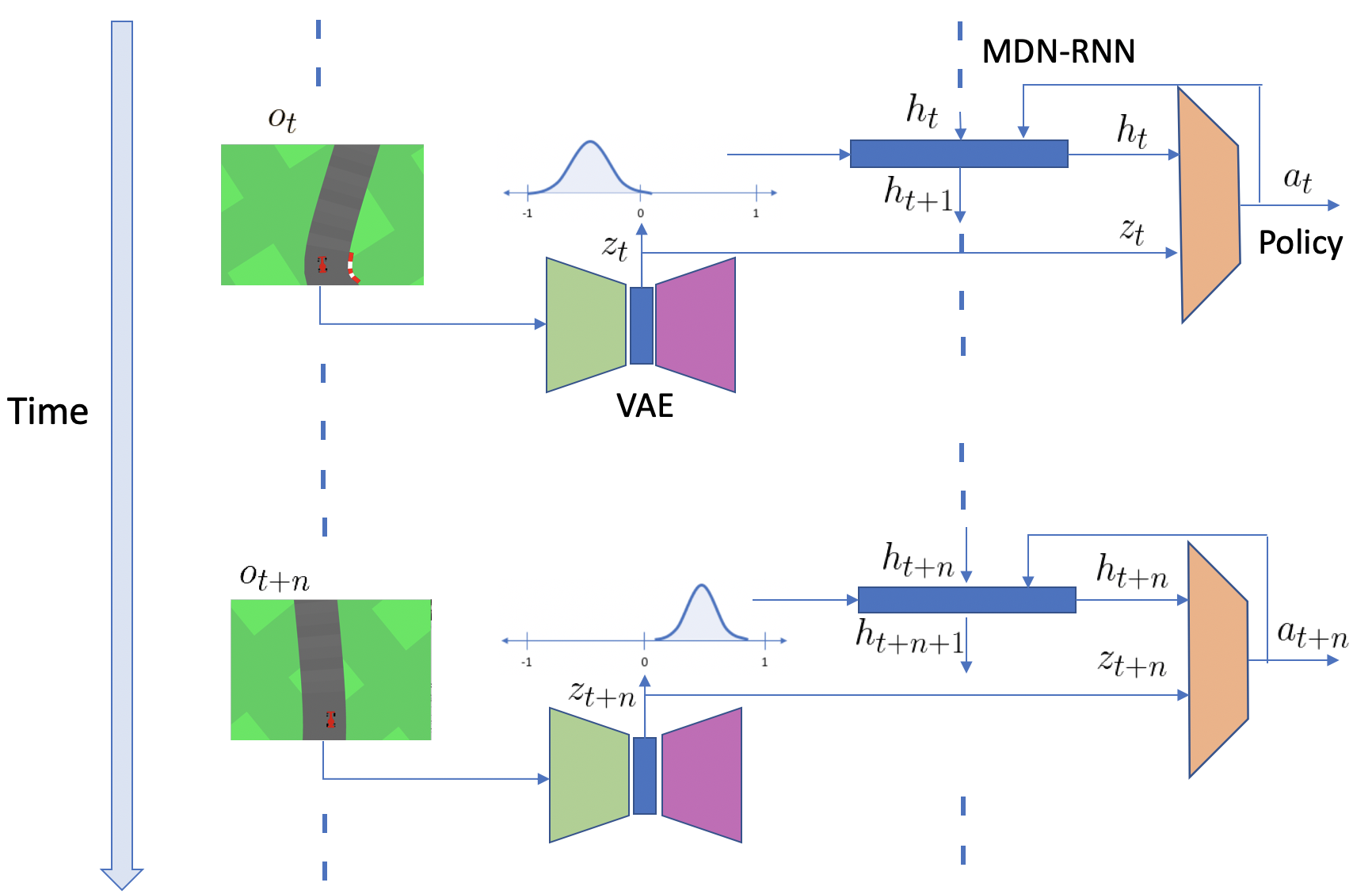}
   \caption{Pictorial representation of world model in action}
   \label{fig:model}
\end{figure}
\end{center}

\section{Background and Related Work}
\label{sec:relatedwork}

\textit{Direct Policy Search} is a class of policy estimation algorithms which find the parameters of the model by optimizing a predefined cost function, which can be with respect to a reward function or expert demonstrations. For a single task, methods on the aspect of Learning a policy from expert demonstrations have been seen to be predominant in the past \cite{DBLP:journals/corr/HesterVPLSPSDOA17, DBLP:conf/atal/ChernovaV07, DBLP:conf/aaai/HesterVPLSPHQSO18, DBLP:conf/icml/KangJF18}. For more complicated tasks involving non-stationary data distribution \cite{DBLP:conf/icra/EnglertPPD13}, methods involve gathering the expert data \cite{DBLP:journals/jmlr/RossGB11}, and training a predictive model. Recently, a lot of methods  as outlined by \cite{DBLP:journals/corr/abs-1801-06503} were proposed in the aspect of model fitting on the expert data for an agent. These methods mainly deal with mitigating, covariate shift, where the input distribution or the training data changes, but the conditional distribution of labels given the data remains fixed \cite{Quionero-Candela:2009:DSM:1462129}. On the other hand, several other works address this problem from an other angle, \cite{DBLP:journals/corr/abs-1801-06503, DBLP:conf/corl/LaskeyLFDG17, DBLP:conf/icml/SunVGBB17}, especially DAGGER \cite{DBLP:journals/jmlr/RossGB11} using active learning \cite{DBLP:conf/nips/HeDE12}. Out of all, we choose DAGGER to train our model as it simple and works well in practical cases. A number of works have been applied to the task of navigation, like \cite{DBLP:conf/icra/CodevillaMLKD18, DBLP:conf/rss/PanCSLYTB18}

\begin{figure}
    \centering
   \includegraphics[width=\columnwidth]{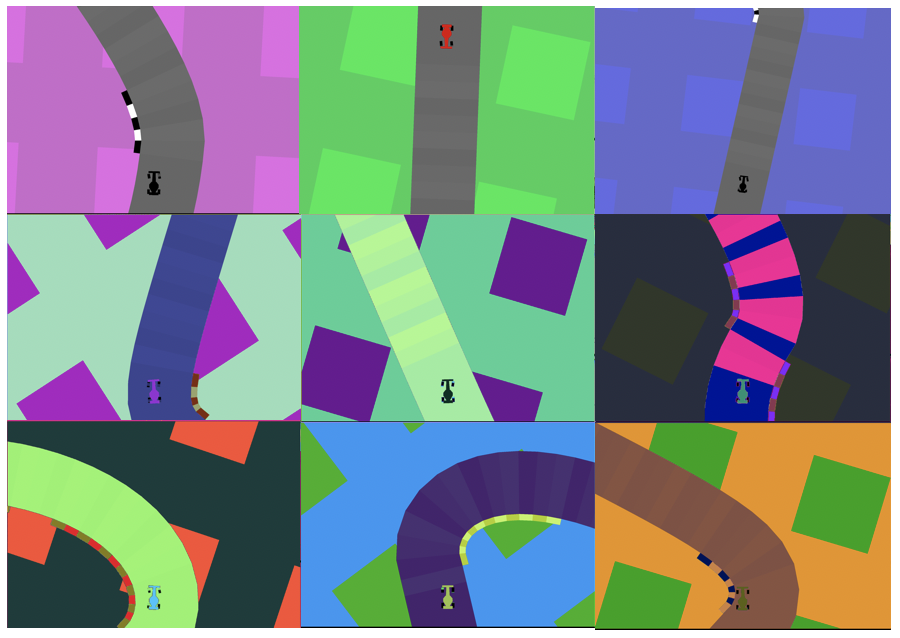}
   \caption{Pictures of different car-racing simulators used. Environments in the first row are used for training tasks, and all the other environments were used as test tasks.}
   \label{fig:realworld}
\end{figure}

Some of the significant improvements in Imitation learning involve making algorithms robust to long-horizon tasks or changing data distribution \cite{DBLP:journals/ftrob/OsaPNBA018, DBLP:conf/icml/0001JADYD18}. Other works, which follow a similar trend, include partitioning the domain into individual tasks and making the model train using multiple tasks \cite{DBLP:conf/globecom/XuLGKAW18}. An enhancement of multi-task imitation learning, i.e., hierarchical imitation learning, involves high-level planners to estimate sub-goals for the low-level policies \cite{DBLP:journals/corr/abs-1810-11043}. Although these methods perform better than naive Imitation learning, most of them do not generalize to new related tasks.


 Many recent works on few-shot imitation learning \cite{DBLP:conf/iclr/0004DLAL17, DBLP:conf/nips/DuanASHSSAZ17, DBLP:conf/corl/FinnYZAL17} involve novel meta-learning schemes. The basic principle underlying all these works involve adapting and inferring model to unseen tasks. Some of the novel approaches used by these methods are hybrid loss functions \cite{DBLP:conf/nips/DuanASHSSAZ17}, evolving policy gradients \cite{DBLP:journals/corr/abs-1802-04821}, and estimating meta parameters \cite{DBLP:conf/icml/FinnAL17}. The core idea of most of the works mentioned above involve parameter adaption for unseen tasks. Some of the few recent works which apply meta-learning to Visual Navigation are \cite{DBLP:journals/corr/SallabSTA17, DBLP:journals/corr/abs-1812-00971}. Compared to others, the meta objective of our adaptive approach relies on the alignment of the evaluated gradients on training data to the test data. Previously, variants of these approaches were used in supervised learning, for minimizing distribution shift between training and test datasets \cite{DBLP:conf/icml/RenZYU18}. Our main contributions are outlined as follows:

1$\cdot$ We propose a novel Importance Weighting method to amplify the gradients evaluated on the training demonstrations for better performance on the test task.

2$\cdot$ Our method is robust on dynamically changing distributions, and can also be extended to Meta Imitation learning, where an agent needs to quickly learn an unseen related task from prior experiences.
    
3$\cdot$ To the best of our knowledge, we are the first to apply iteratively trained world models, along with our proposed improvements, to the task of Imitation learning

This paper has been organized as follows. In Section \hypersetup{linkcolor=blue}\ref{sec:worldmodel}, World models and our improvements are illustrated. Then, In Section \hypersetup{linkcolor=blue}\ref{sec:method}, out main contribution is outlined. After that, different experiments performed and the related observations are explained in \hypersetup{linkcolor=blue}\ref{sec:experimentalsetup} and \hypersetup{linkcolor=blue}\ref{sec:observation} respectively, followed by conclusion.

\begin{figure}
    \centering
   \includegraphics[width=\columnwidth]{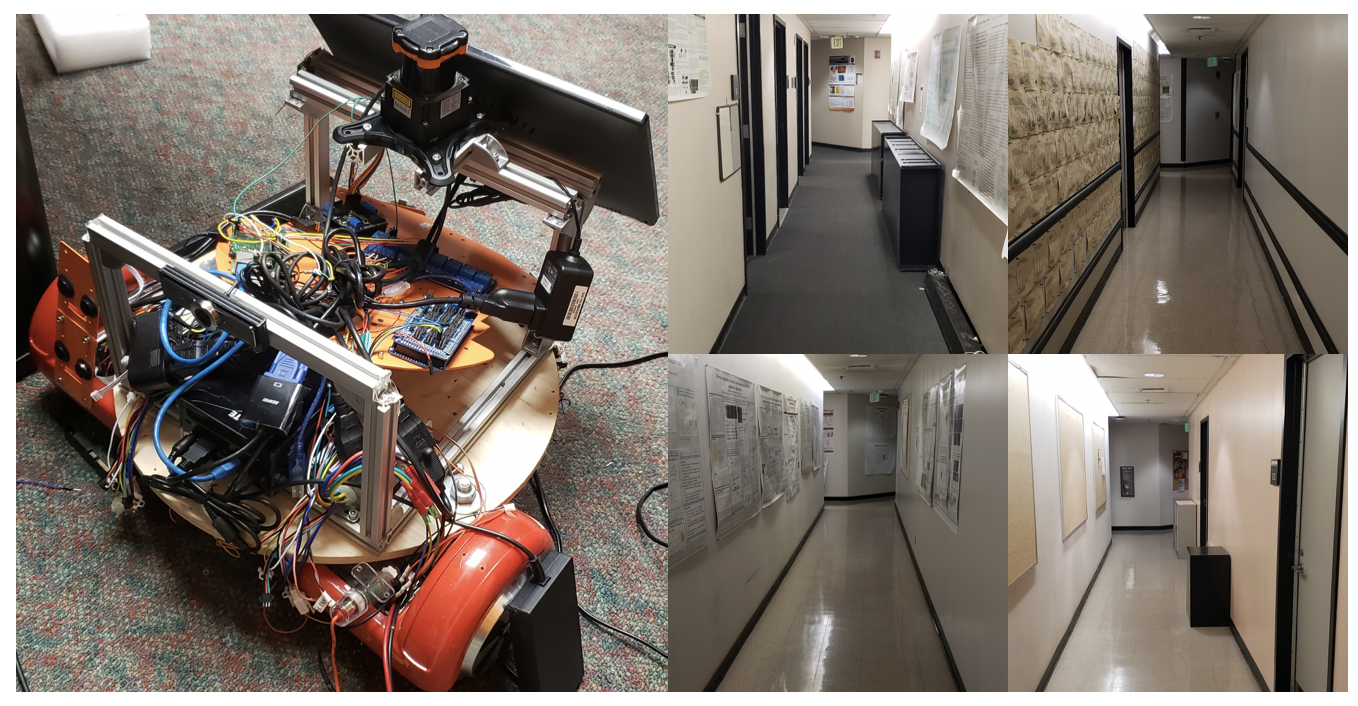}
   \caption{Picture of the physical robot used to test our method. On the right side, the real-world environments used for evaluating our model are pictured.}
   \label{fig:realworld}
\end{figure}

\section{Imitation learning using World models}
\label{sec:worldmodel}

Instead of learning the parameters by training the model end to end like \cite{DBLP:conf/icra/CodevillaMLKD18}, an Itervative training procedure using World models \cite{DBLP:conf/nips/HaS18} is adapted. We hypothesize that modularizing a model and training each component individually works well for many tasks. We corroborate this hypothesis with previous works in Neuroscience such as \cite{DBLP:journals/tamd/Schmidhuber10, Schmidhuber1990}. This methodology also performs well for meta-learning scenarios, where the end policy alone can be retrained for new tasks by leaving prior modules fixed \footnote{Prior modules can also be adapted just as the end policy. We leave this for our future work.}, as it will be clear in the later sections.

\begin{figure*}
  \includegraphics[width=\textwidth]{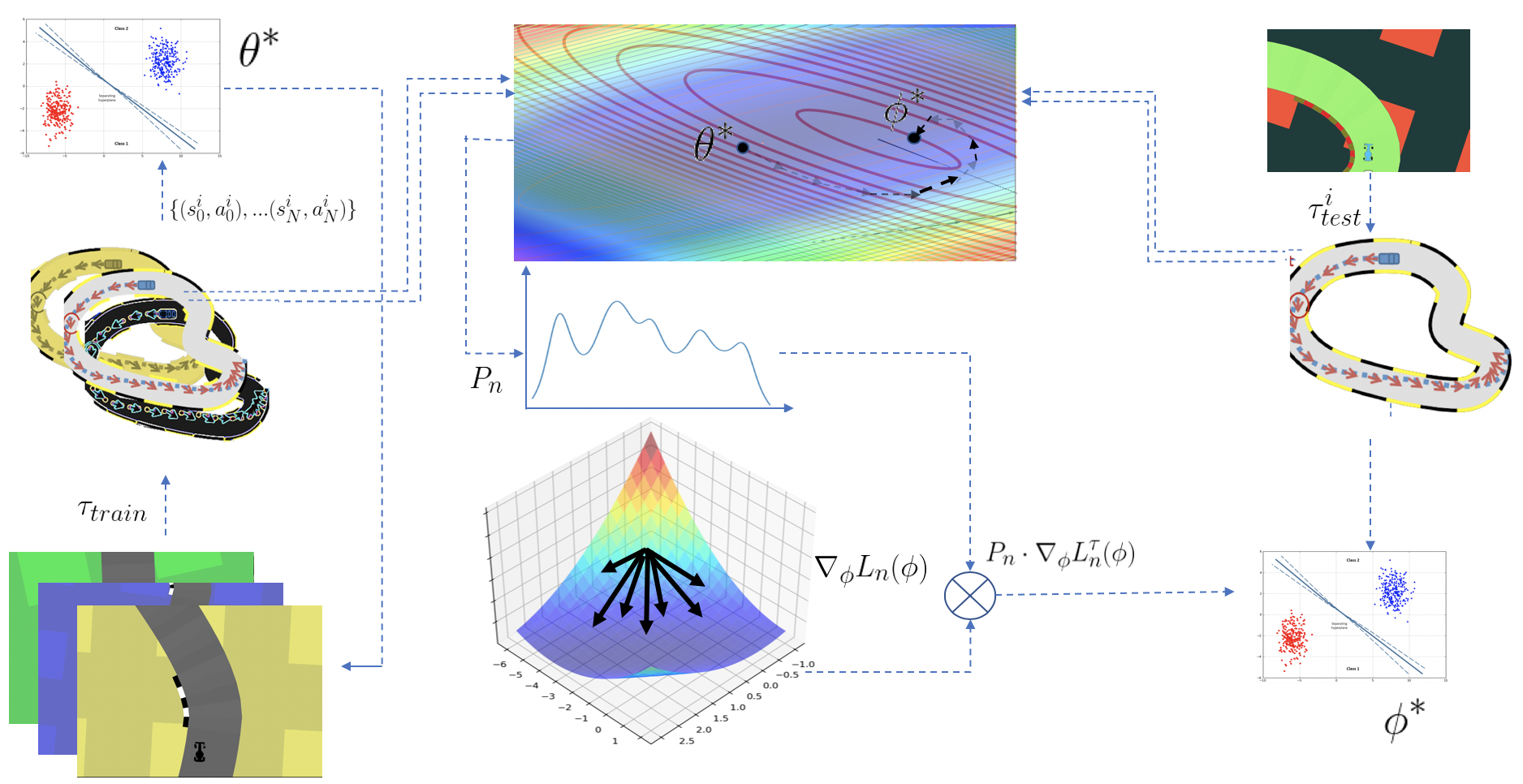}
  \caption{Working of our method. On the left side, the training procedure is mentioned. The agent is trained on a variety of environments. During test time, our method enables the agent has to adapt to a new environment.}
  \label{fig:distribution}
\end{figure*}

\subsection{Prelimnaries}
 A World model (Figure \ref{fig:model}) consists of a Variational Autoencoder (VAE), Mixture Density Recurrent Neural Network (MDN-RNN) and a Neural Network (Policy). Each of these modules are named as Vision (V), Memory (M) and Controller (C) modules. Note that in the remainder of the paper, we will be using policy and controller interchangeably.
 
 When a world model is evaluated on a specific task, say $\mathcal{T}^i$, we obtain a set of observation-action pairs $\{\langle O^i_j$, $a^i_j \rangle \}_{j=0}^{j=T}$. During training, A VAE, encodes an observation $O_i^j$ to a latent variable $z_j$ which is sampled from a Normal distribution $z \in \mathcal{N}(\mu, \Sigma)$. In tasks like navigation where the current state is conditional on the previous state and action, we have a transition model (MDN-RNN) which models $p(z_{j+1} | z_{j}, a_j)$. Specifically, MDN-RNN emits $h_t$ at every time-step, which contains the transition probabilities. A state is a vector formed by concatenating $h_j$ and $z_j$. In the case of Model-free approaches, $s_j$ would just be $z_j$. Given this state, a policy is a single layer neural network $f(s_j, \theta)$ 
 

\subsection{Improvements on existing World Models}
To train the V and M modules, Instead of spawning trajectories with a random policy \cite{DBLP:conf/nips/HaS18}, we use a trained policy to collect data. We performed some experiments, where we found that reproducing the original World model experiments on an  V and M trained on data from an expert policy, resulted in better performance. We also found that World models generalize very well, even with small amounts of data trajectories, (of the order 10-20), which makes them suitable for applying in real-world settings.

\begin{figure*}
  \includegraphics[width=\textwidth]{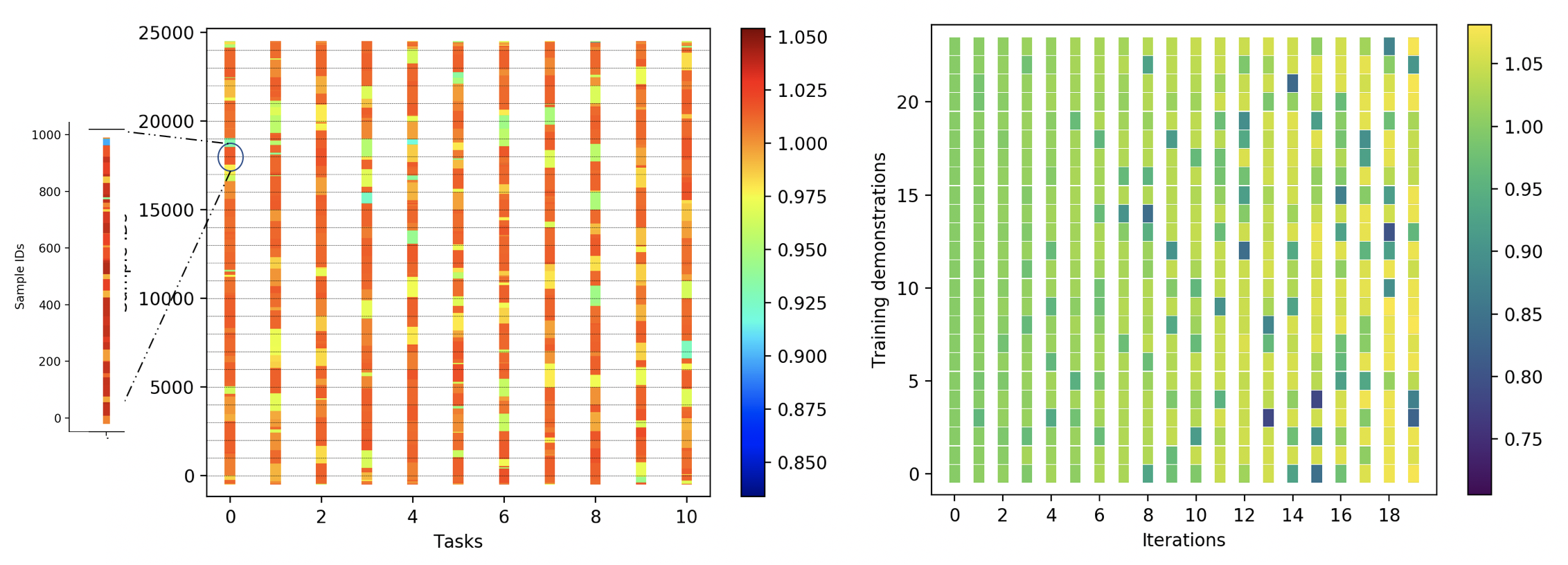}
  \caption{(a) Colormap of $p$ values after $\phi^i$ converges on 10 different tasks. The dashed line represents the trajectories. An instance of the $p$ values of a trajectory is enlarged for better visualization. (b) Change of $p$ values over time. We evaluated our algorithm on a specific task to show how the $p$ values change over every iteration.}
  \label{fig:distribution}
\end{figure*}


\subsection{Active learning for policy estimation} 
In this work, we train world models using Imitation learning by employing DAGGER as proposed by \cite{DBLP:journals/jmlr/RossGB11}. DAGGER is an active learning-based method which involves giving control to the agent to gather training data. The policy is then trained by aggregating data over each trial on the environment. We used the original world model \cite{DBLP:conf/nips/HaS18} as an expert world model and our improved world model as the agent's model. The policy parameters of the original world model were trained using Reinforcement learning and performed particularly well, even amidst any disturbances or noise, in the environment. This specific trait made it the right choice for an expert in this case, as in many situations, the agent would go to a vulnerable state, and the expert should recover from such erratic behavior. In the case of the real-world robot, we do not have such a robust expert yet, and so we used a human demonstrator.

\section{Proposed method}
The main contribution is outlined. The code related to our method can be found at \url{https://github.com/kiran4399/weighted_learning}.\\

\label{sec:method}
\subsection{Problem formulation}
We consider a Meta-Supervised learning setting, where the agent has access to the distribution of training tasks $p(\mathcal{T}_{train})$ at train-time. To train the policy, we require a set of expert demonstrations $\{\langle s^i_j, \hat{a}^i_j\rangle\}_{j=0}^{j=T}$, where the $T$ is the length of an episode, on the set of train tasks $\mathcal{T}^i_{train}$ to estimate $\theta^*$ which minimizes the log-likelihood, on the expert data. The goal of the agent is to generalize to new test task $\mathcal{T}^i_{test}$ using a few samples. The train and test datasets consist of the data aggregated on the training tasks and the test specific task, respectively. $\theta^*$ is $\theta$ after convergence on the training dataset and, is used to initialize $\phi^i$, which are the policy parameters required to train on the test data obtained on a specific task.

\subsection{Assigning Importance weights}
In general, Imitation learning is based on optimizing a model maximize the log-likelihood which is represented in the following equation, where $f$ is the model and $N$, $s_n$, $\hat{a}_n$ are the number of samples, states and actions respectively.

\begin{equation}
 \mathcal{L}(\theta) = \frac{1}{N}\sum_{n=0}^{n=N} \hat{a}_n\log(f(s_{n}))
\end{equation}
\begin{equation}
    \theta^* = \argmax_{\theta} \frac{1}{N}\sum_{n=0}^{n=N} \mathcal{L}(\theta)
\end{equation}

However, as mentioned before, maximizing the average likelihood may not yield the desired outcome in a lot of applications because of some training samples being irrelevant, noisy or unevenly distributed. We can correct the covariate shift by estimating a non-trivial distribution of scalar weights, estimated from a small data batch drawn from an optimal distribution \cite{DBLP:conf/nips/HaS18}. Compared to Eqn 2, the optimal parameters and the gradient of the Loss function with respect to parameters $\phi$ becomes:

\begin{equation}
    \phi^* = \argmax_{\phi} \sum_{n=0}^{n=N} P_n \mathcal{L}(\phi)
\end{equation}

\begin{equation}
\begin{split}
\nabla_{\phi} \mathcal{L}(\phi) &= \sum_{n=0}^{n=N} P_n \cdot \nabla_{\phi} \mathcal{L}_n(\phi)
\end{split}
\end{equation}

In the above equation, $P$ is a vector of the size of training samples and adapts based on the training and test data. For meta-imitation learning, we can use same method to learn the task distribution shift and make the training adapt to different perturbed scenarios. In other words, during test-time, we can evaluate gradients $\nabla_{\phi} L(\phi)$ on task-specific demonstrations $\mathcal{D}_{test}$ and impose them on the per-sample gradients $\nabla_{\phi} \mathcal{L}_{n}(\phi)$ estimated on the training data $\mathcal{D}_{train}$, where $\phi^i$ are the policy parameters.

Initially, we train the policy parameters $\theta$ on a sampled set of train tasks $\mathcal{T}^i_{train}$ to collect training data Note that, $\theta$ and $\phi$ are the parameters of the policy at train and test time respectively. We could have used $\theta^i$, instead of $\phi^i$, but we use that notation for parameters which are obtained by training the policy only on the test task $\mathcal{T}_{test}^i$.

During test time, when we asses the generalization ability of a policy on an unseen test task, we train the policy likewise to the training approach, but only using the test data. We, however, use the test data to learn the $P$ distribution and calculate the dot product, represented by $\sbullet$ with the per sample gradients, which we use to update $\phi$. For simplicity sake superscript $i$ is omitted. The distribution $P$, which is a vector of size $N$, can be updated by optimizing the cost function $J(P, \phi)$ of the L2 distance between $\nabla_{\phi} L(\phi)$ and $\nabla_{\phi} \mathcal{L}(\phi)$ for each batch of test and train data. Note that if the dimensionality of the average gradients  $\nabla_{\phi} L(\phi)$ is $\mathcal{R}^{a \times b}$, the dimensions of $\nabla_{\phi} \mathcal{L}(\phi)$ would be $\mathcal{R}^{N \times a \times b}$ Also, we apply softmax over distribution parameters $p_n$  such that $P_n$ always sum to 1. The gradient of the cost function with respect to $P$ is as follows.

\begin{equation}
\frac{\partial}{\partial P}J(P, \phi) = \frac{\partial}{\partial P} \left(P \sbullet \nabla_{\phi} \mathcal{L}_n(\phi) - \nabla_{\phi} \hat{\mathcal{L}}(\phi)\right)^2
\end{equation}

\begin{equation}
    \nabla_P J(P, \phi) = \left[\nabla_{\phi} L_n(\phi)^T \sbullet \left[ P_n \sbullet \nabla_{\phi} \mathcal{L}_n(\phi) - \nabla_{\phi^{'}} L(\phi)\right]\right]^T
\end{equation}

During every iteration, apart from updating $\phi$ using our method, we also update the distribution parameters $p_n$, for $K$ iterations. Using the analytical gradient computed in Eqn 6, we can compute the gradient with respect to $p$ using chain rule as follows.
    
\begin{equation}
    \frac{\partial J}{\partial p} = \frac{\partial J}{\partial P} \sbullet \frac{\partial P}{\partial p}
\end{equation}
\begin{equation}
    \nabla_p J(p, \phi) = \nabla_P J(P, \phi) \sbullet \nabla_p \sigma(p)
\end{equation}

In the above equation $\nabla_p \sigma(p)$ is the gradient of the softmax function, which is a $N \times N$ matrix. The policy parameters $\phi^i$ are thus, iteratively learned by utilizing the training data, but deriving the $P$ distribution from the test data. A synopsis of the entire algorithm is illustrated on the next page.

\begin{algorithm}
\caption{Estimate $\theta^*$ and $\phi^{i^*}$}
\small
\begin{algorithmic}[1]
\REQUIRE $p(\mathcal{T}_{train})$ and $p(\mathcal{T}_{test})$ as task distributions
\REQUIRE $\alpha$, $\beta$ and $\gamma$ as step-size parameters 
\REQUIRE Expert policy $\hat{\theta}$
\STATE Sample tasks $\mathcal{T}^i$ from task distribution $p(\mathcal{T}_{train})$
\STATE Evaluate $\hat{\theta}$ on {$\mathcal{T}^1, \mathcal{T}^2, .., \mathcal{T}^{M}$} to collect \{$O_j^i$, $\hat{a}_j^i$\}
\STATE Train VAE and MDN-RNN (Refer Section \ref{sec:worldmodel})

\STATE Randomly initialize $\theta$

\FOR{each sampled task $\mathcal{T}^{i}$ from $p(\mathcal{T}_{train})$}
\STATE Evaluate $\theta_{\tau}$ on $\mathcal{T}^j_{train}$ to generate $\mathcal{D}^i_{train}$
\STATE $\mathcal{D}_{train} \leftarrow \mathcal{D}_{train} \cup \mathcal{D}^i_{train}$
\FOR{$t \in \{0, 1, .. \tau$ \}}
\STATE Evaluate gradients on $\mathcal{D}_{train}$ at $\theta_{t}$
\STATE $\theta_{t+1} \leftarrow \theta_{t} - \alpha\nabla_{\theta}L(\theta_{t})$
\ENDFOR
\ENDFOR

\STATE Sample tasks $\mathcal{T}^{i}$ from $p(\mathcal{T}_{test})$
\FOR{each task $\mathcal{T}^{i}$}
\STATE Initialize $\phi^i$ = $\theta^*$
\FOR{$t$ in \{0, 1, .. $\hat{\tau}$ \}}

\STATE Evaluate $\phi^i_t$ on $\tau^i_{test}$ to generate $\mathcal{D}_{test}$

\FOR{$k \in {0, 1, ... K}$}
\STATE Evaluate $\nabla_{\phi} L(\phi_{t})$ on $D^i_{test}$
\STATE Calculate $\nabla_P J(P, \phi)$ From Eq 8.
\STATE $\nabla_p J(p, \phi^i) = \nabla_P J(P, \phi^i) \sbullet \nabla \sigma$
\STATE $p_{k+1} \leftarrow p_{k} - \gamma\nabla_{p} J(p, \phi)$
\ENDFOR

\STATE $\phi^i_t \leftarrow \phi^i_t - \beta\nabla_{\phi} \hat{L}(\phi_t)$
\ENDFOR
\ENDFOR
\end{algorithmic}
\end{algorithm}

\section{Experimental setup}
\label{sec:experimentalsetup}
\subsection{Experiments on simulator}
We used a Car Racing simulator from OpenAI gym to test our method. We adapted a world-model from \cite{DBLP:conf/nips/HaS18} by retraining the architectures for the VAE and MDN-RNN, using different hyperparameters. The controller/policy was changed to a single layer classifier to categorize a state to one of the five discretized actions. We chose to use a single layer policy, as simple architectures tend to generalize better. We created multiple car racing environments and considered them as tasks. We collect 24 expert trajectories, 4 for each task, and use them as the training data.


For policy training, we used Stochastic Gradient Descent (SGD) with step-size parameters $\alpha$, $\beta$ and $\gamma$ as $0.01$, $0.01$ and $0.05$ respectively. We limited $K$ updates to 10, as we found this to be sufficient for the experiments. We set $\tau$ to 3000 and 4000 iterations for training and testing respectively. We encourage the reader to refer to the algorithm in the previous page for the notations.

\subsection{Physical setup}
We also evaluated our method using a physical system, in our case a Non-Holonomic, differential-drive based robot for the task of visual navigation. For real-world experiments, we defined a task as the environment on a specific level in the building. Each level was visually very distinct from the other, and we interpreted an environmental seed as a unique source and destination location on a particular level of the building. Images obtained from a camera, mounted on the robot, were timestamped and synced with the action commands before being sent for training. We used ROS and Tensorflow for implementing our experiments.

We used a world model with similar modifications applied to the simulator experiments. We collected 30 human-controlled trajectories, 10 for each task, and trained the V and M modules. The robot was then tested on the remaining one environment, which it had not seen before.

\section{Observations and Results}
\label{sec:observation}

Compared to the state of the art benchmark \cite{DBLP:conf/nips/HaS18} on the car-racing simulator, our method deals not so much with the maximum score in a given episode but how quickly it can learn an optimal behavior and adapt to a related unseen environment. Following are some of the observations, which we had found.

\subsection{Generalization to unseen Tasks}
\begin{figure}
    \centering
    \includegraphics[width=\columnwidth]{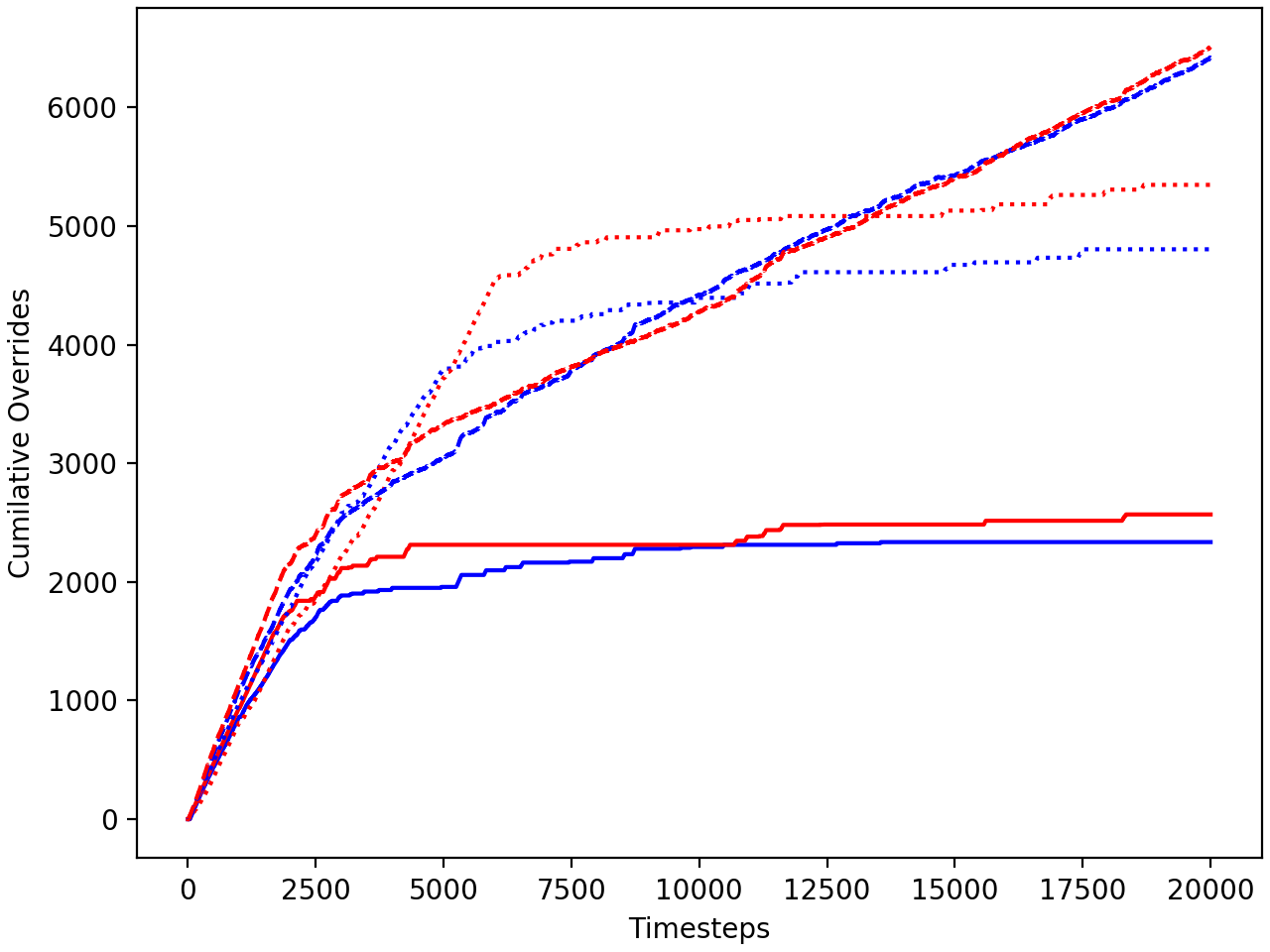}
    \caption{Comparision of 2 different tasks (marked with red and blue) on which the model was evaluated. The plain, dotted and dashed curve represents the performance of $\theta^*$, $\theta^{i^*}$ and $\phi^{i^*}$ baselines respectively. We can see that our method outperforms other baselines. The $y$ axis represents the cumulative number of overrides over the trials, and the $x$ axis represents timesteps of all the trials combined.}
    \label{fig:cumover}
\end{figure}

Apart from the training data, We also generated some environments as test tasks for the model. Though none of the components of the world model were trained on those tasks, our method made the world model generalized to them. We also added uniform Gaussian noise on all the observations of the test tasks for robustness. To compare our method, we used 2 baselines: DAGGER baseline ($\theta^*$) which was trained by aggregating the data from all the prior tasks and the test task and Fine-tuning baseline ($\theta^{i^*}$) which was trained the policy only on the test task. Figure \hypersetup{linkcolor=red}\ref{fig:cumover} shows that our method outperformed these baselines. As a primary measure, we used the number of times the expert had to intervene to allow the agent to get off a vulnerable state, which we call \textit{override}. We also portrayed the mean accuracies on different tasks for each baseline, as we wanted the reader to notice the relationship between accuracy and overrides as an appropriate measure for comparing baselines. We argue that both of them are required in active imitation learning, as a model might have a high average accuracy, but might not perform well on some important states. For quantitative comparison on different baselines, refer Table \hypersetup{linkcolor=red}\ref{tab:results}

\subsection{Converging to local optima for train tasks}
Usually, in Meta-learning, the goal for the classifier is the generalize to unseen tasks from the prior information obtained from the train tasks. However, in some scenarios, like that of navigation, we want the agent to perform well on the training tasks as well. After $\theta$ convergence, we evaluate the policy on each training tasks, sampled with random seeds and added Gaussian noise. Surprisingly, the agent performed sub-optimally on every task. Since the world model was trained on the training tasks, Naively aggregated data should've performed well. However, in situations, where there are a sufficiently large number of tasks, the model would collapse to a local optimum. However, when we ran used our method for on a specific train task as a test task for 1 iteration. it resulted in better performance.

\subsection{Robustness amongst noise in demonstrations}
\begin{figure}[htp]
  \centering
  \subfigure{\includegraphics[scale=0.25]{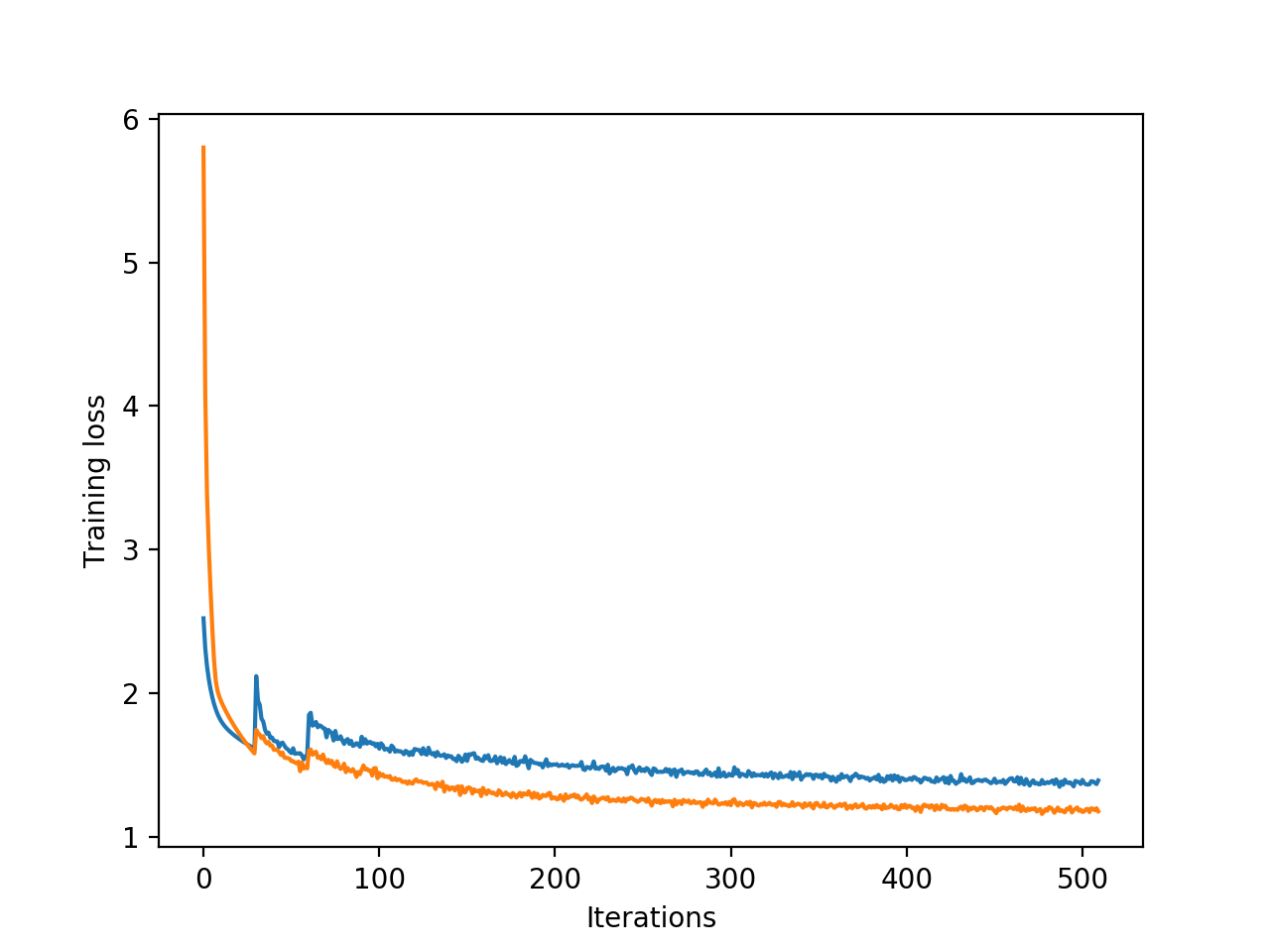}}
  \subfigure{\includegraphics[scale=0.25]{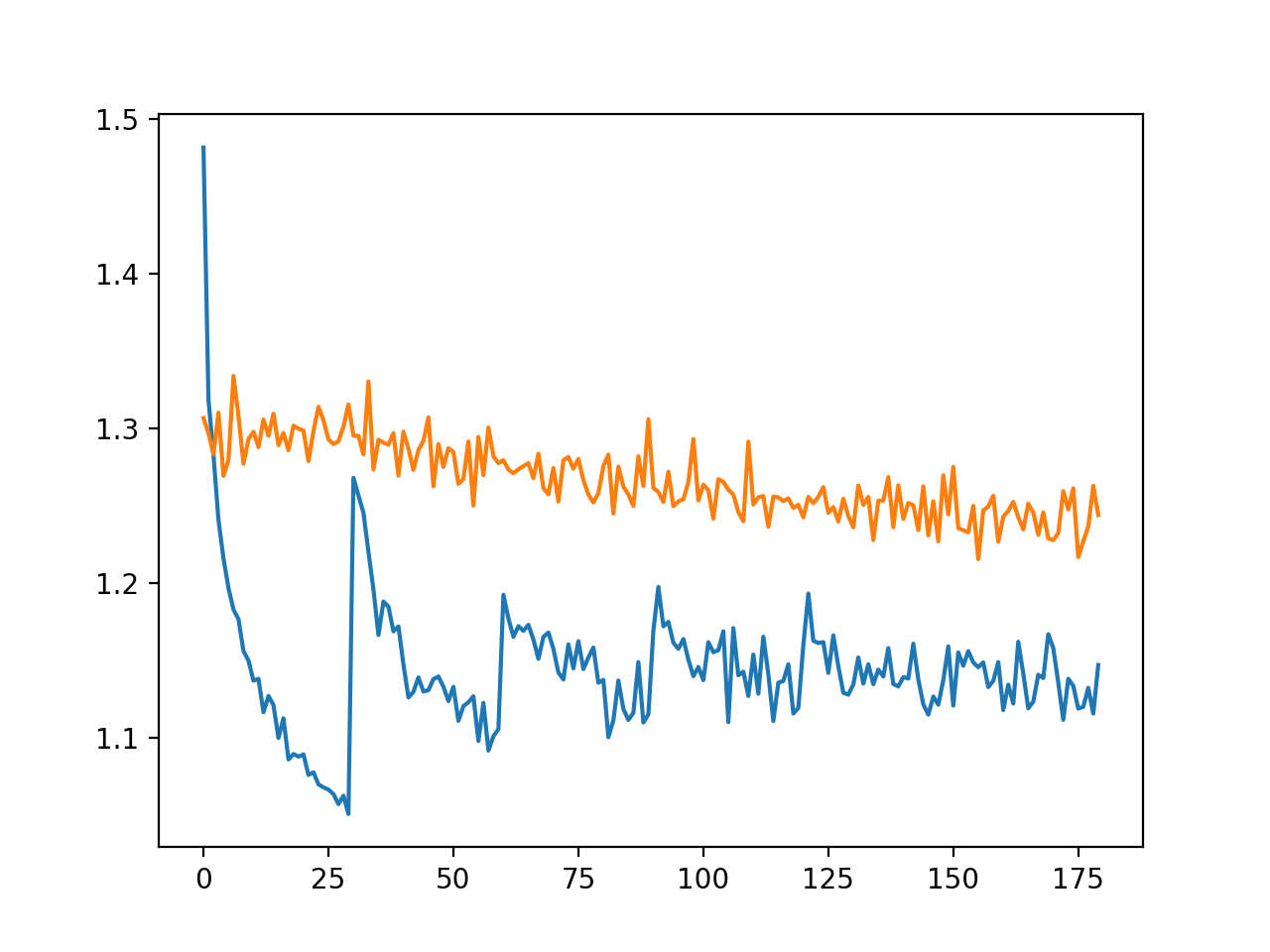}}
  \subfigure{\includegraphics[scale=0.25]{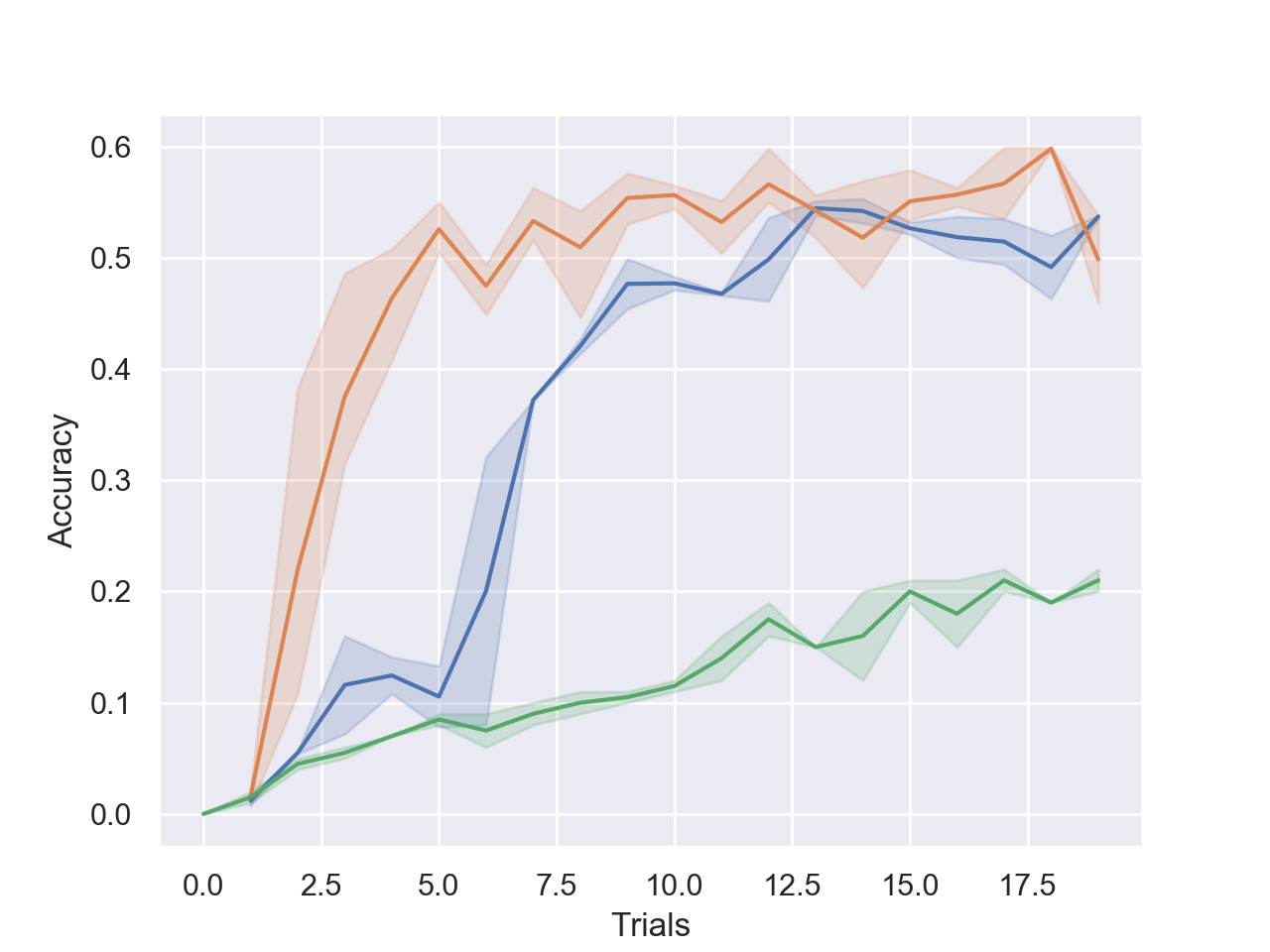}}
  \subfigure{\includegraphics[scale=0.25]{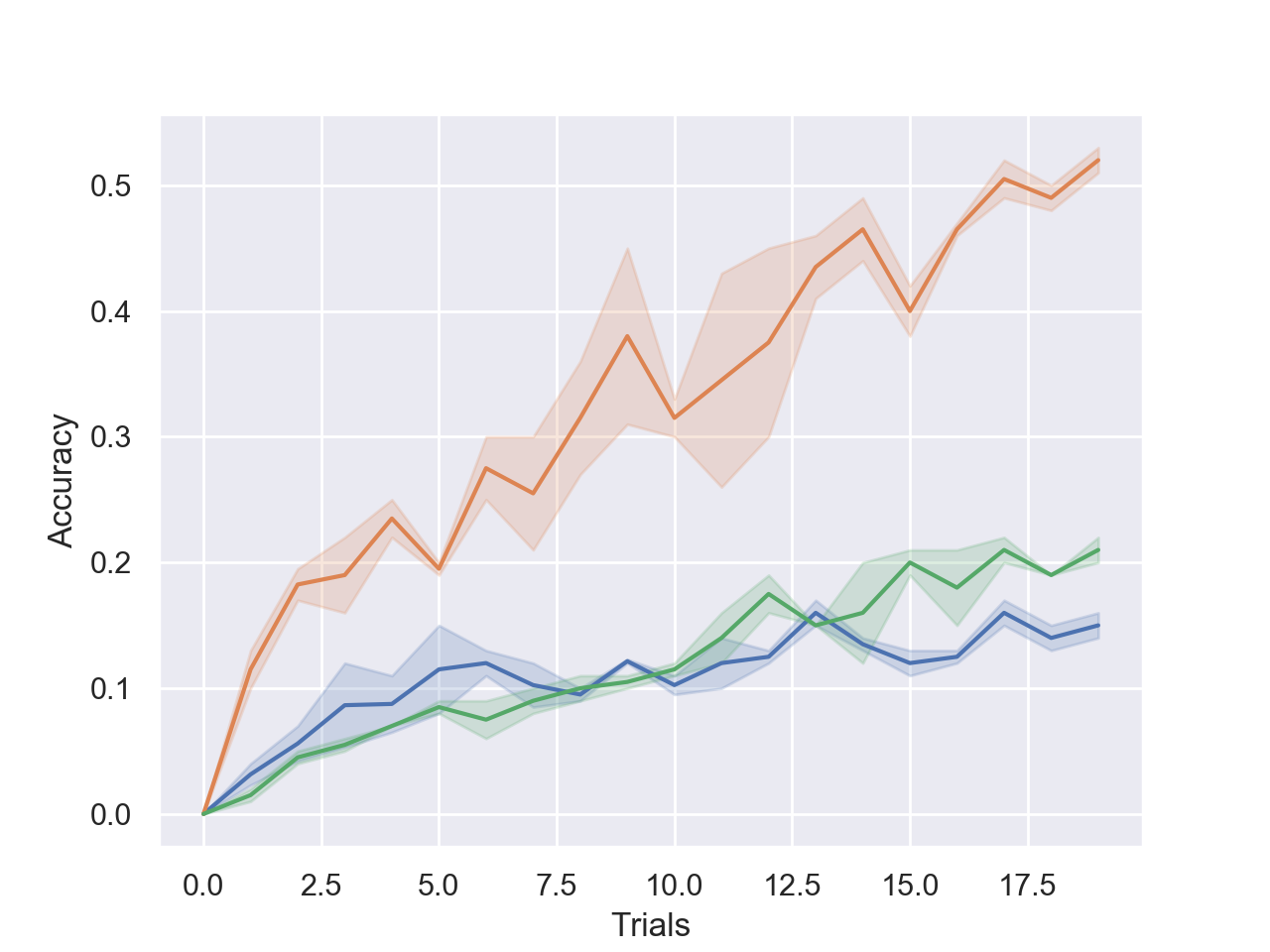}}
  \caption{Plots corresponding to (a) Comparison of training loss with time for $\theta^*$ (orange) and $\theta^{i^*}$ (blue) during training (b) Comparision of training loss on training data and test data of $\phi^{i^*}$ on a randomly sampled Unseen task. (c) Comparison of the Accuracy of $\theta^*$ (green), $\theta^{i^*}$ (blue) and $\phi^*$ (orange) on a Clean dataset (lower left) Noisy dataset (lower right), for a set of 4 unseen tasks.}
  \label{fig:plots}
\end{figure}

Our algorithm works well, even in cases where there is noise in the collected demonstrations. During training, in each iteration of policy evaluation, we randomly select 50 \% of the collected demonstrations and corrupt the action labels. Even in such scenarios, our algorithm remains robust by giving those corrupted samples less importance, i.e., less $p$ values and performing well on the test task. Figure \hypersetup{linkcolor=red}\ref{fig:plots} states the results.

\subsection{Visualization of the P values}
Although we performed many experiments confirming the robustness of our algorithm, it is important to understand the $p$ values in each case, to know which of the samples are getting more importance. Note that, each aggregate training sample, which are in total 24k, has a $p$ value associated with it. In Figure \hypersetup{linkcolor=red}\ref{fig:distribution}, we provide a color map of the $p$ values, after $\phi^i$ converges. Although it is unclear, how the samples having high $p$ are related to the test task, we can comprehend that the model is able to learn the new task from the prior tasks. In the same figure, we also show how the $p$ values change with each trial, which indicates that, the sample importance, changes with getting more information from the new test task.

\subsection{Experiments in real-world}
We ran our robot on the corridors in the Hedco Neuroscience building at the University of Southern California. Though the geographical and structural maps of each floor were similar, the visual features were very different, which makes it perfect for applying our method. Pictures taken in different floors are depicted in Figure \hypersetup{linkcolor=red}\ref{fig:realworld}. Results obtained by test the robot in different environments are shown in Table \hypersetup{linkcolor=red}\ref{tab:results}

\begin{table}[ht]
\small
\centering
\caption{Quantitative of the baselines. The results given in the following 2 tables correspond to the Car Racing simulator and the real world environments respectively. The values mentioned in each cell correspond to the accuracies of the policy after training on the number of trials.}

\begin{tabular}{|m{1.4cm}|m{1.7cm}|m{1.7cm}|m{1.7cm}|}
\hline
Method & 1 Trial & 2 Trials & 5 Trails \\ \hline \hline
DAGGER & $5.1\pm0.8\%$ & $6.2\pm1.3\%$ & $8.5\pm1.1\%$ \\
\hline
Fine-tune & $9.2\pm0.9\%$ & $12.6\pm1.5\%$ & $17.9\pm2.8\%$ \\
\hline
\textbf{Ours} & $\mathbf{21.5\pm5.1\%}$ & $\mathbf{39.7\pm7.2\%}$ & $\mathbf{53.2\pm3.3\%}$ \\
\hline
\end{tabular}

\vspace*{0.5 cm}

\begin{tabular}{|m{1.4cm}|m{1.7cm}|m{1.7cm}|m{1.7cm}|}
\hline
Method & 2 Trial & 5 Trials & 10 Trails \\ \hline \hline
DAGGER & $3.5\%$ & $7.1\%$ & $7.9\%$ \\
\hline
Fine-tune & $8.1\%$ & $9.3\%$ & $13.1\%$ \\
\hline
\textbf{Ours} & $\mathbf{18.2\%}$ & $\mathbf{23.5\%}$ & $\mathbf{31.6\%}$ \\
\hline
\end{tabular}
\label{tab:results}
\end{table}%

\section{Conclusion}
\label{sec:discussion}
In this paper, we presented a Meta-Imitation learning algorithm which involves learning new skills from prior knowledge. We defined a task or skill as an environment having a specific data distribution attributed by time or situation. applications, which involves substantial covariate shifts, by considering it as a meta-learning problem. We have also shown how the proposed algorithm can be used to improve the policy performance on a single task, which was trained on a set of tasks Some of our experiments performed using a real robot, shows how our algorithm can aid real-world scenarios as well. The results shown on the task of navigation support our assertions.



\printbibliography

@inproceedings{pathakICMl17curiosity,
    Author = {Pathak, Deepak and Agrawal, Pulkit and
              Efros, Alexei A. and Darrell, Trevor},
    Title = {Curiosity-driven Exploration by Self-supervised Prediction},
    Booktitle = {International Conference on Machine Learning ({ICML})},
    Year = {2017}
}

@inproceedings{DBLP:conf/aaai/HesterVPLSPHQSO18,
  author    = {Todd Hester and
               Matej Vecerik and
               Olivier Pietquin and
               Marc Lanctot and
               Tom Schaul and
               Bilal Piot and
               Dan Horgan and
               John Quan and
               Andrew Sendonaris and
               Ian Osband and
               Gabriel Dulac{-}Arnold and
               John Agapiou and
               Joel Z. Leibo and
               Audrunas Gruslys},
  title     = {Deep Q-learning From Demonstrations},
  booktitle = {Proceedings of the Thirty-Second {AAAI} Conference on Artificial Intelligence,
               (AAAI-18), the 30th innovative Applications of Artificial Intelligence
               (IAAI-18), and the 8th {AAAI} Symposium on Educational Advances in
               Artificial Intelligence (EAAI-18), New Orleans, Louisiana, USA, February
               2-7, 2018},
  pages     = {3223--3230},
  year      = {2018},
  crossref  = {DBLP:conf/aaai/2018},
  url       = {https://www.aaai.org/ocs/index.php/AAAI/AAAI18/paper/view/16976},
  bibsource = {dblp computer science bibliography, https://dblp.org}
}

@inproceedings{DBLP:conf/nips/HeDE12,
  author    = {He He and
               Hal Daum{\'{e}} III and
               Jason Eisner},
  title     = {Imitation Learning by Coaching},
  booktitle = {Advances in Neural Information Processing Systems 25: 26th Annual
               Conference on Neural Information Processing Systems 2012. Proceedings
               of a meeting held December 3-6, 2012, Lake Tahoe, Nevada, United States.},
  pages     = {3158--3166},
  year      = {2012},
  crossref  = {DBLP:conf/nips/2012},
  url       = {http://papers.nips.cc/paper/4545-imitation-learning-by-coaching},
  bibsource = {dblp computer science bibliography, https://dblp.org}
}

@inproceedings{DBLP:conf/icml/SunVGBB17,
  author    = {Wen Sun and
               Arun Venkatraman and
               Geoffrey J. Gordon and
               Byron Boots and
               J. Andrew Bagnell},
  title     = {Deeply AggreVaTeD: Differentiable Imitation Learning for Sequential
               Prediction},
  booktitle = {Proceedings of the 34th International Conference on Machine Learning,
               {ICML} 2017, Sydney, NSW, Australia, 6-11 August 2017},
  pages     = {3309--3318},
  year      = {2017},
  crossref  = {DBLP:conf/icml/2017},
  url       = {http://proceedings.mlr.press/v70/sun17d.html},
  bibsource = {dblp computer science bibliography, https://dblp.org}
}

@article{DBLP:journals/corr/abs-1802-04821,
  author    = {Rein Houthooft and
               Richard Y. Chen and
               Phillip Isola and
               Bradly C. Stadie and
               Filip Wolski and
               Jonathan Ho and
               Pieter Abbeel},
  title     = {Evolved Policy Gradients},
  journal   = {CoRR},
  volume    = {abs/1802.04821},
  year      = {2018},
  url       = {http://arxiv.org/abs/1802.04821},
  archivePrefix = {arXiv},
  eprint    = {1802.04821},
  bibsource = {dblp computer science bibliography, https://dblp.org}
}

@inproceedings{DBLP:conf/atal/ChernovaV07,
  author    = {Sonia Chernova and
               Manuela M. Veloso},
  title     = {Confidence-based policy learning from demonstration using Gaussian
               mixture models},
  booktitle = {6th International Joint Conference on Autonomous Agents and Multiagent
               Systems {(AAMAS} 2007), Honolulu, Hawaii, USA, May 14-18, 2007},
  pages     = {233},
  year      = {2007},
  crossref  = {DBLP:conf/atal/2007},
  url       = {https://doi.org/10.1145/1329125.1329407},
  doi       = {10.1145/1329125.1329407},
  bibsource = {dblp computer science bibliography, https://dblp.org}
}

@article{DBLP:journals/corr/HesterVPLSPSDOA17,
  author    = {Todd Hester and
               Matej Vecerik and
               Olivier Pietquin and
               Marc Lanctot and
               Tom Schaul and
               Bilal Piot and
               Andrew Sendonaris and
               Gabriel Dulac{-}Arnold and
               Ian Osband and
               John Agapiou and
               Joel Z. Leibo and
               Audrunas Gruslys},
  title     = {Learning from Demonstrations for Real World Reinforcement Learning},
  journal   = {CoRR},
  volume    = {abs/1704.03732},
  year      = {2017},
  url       = {http://arxiv.org/abs/1704.03732},
  archivePrefix = {arXiv},
  eprint    = {1704.03732},
  bibsource = {dblp computer science bibliography, https://dblp.org}
}

@inproceedings{DBLP:conf/icml/KangJF18,
  author    = {Bingyi Kang and
               Zequn Jie and
               Jiashi Feng},
  title     = {Policy Optimization with Demonstrations},
  booktitle = {Proceedings of the 35th International Conference on Machine Learning,
               {ICML} 2018, Stockholmsm{\"{a}}ssan, Stockholm, Sweden, July
               10-15, 2018},
  pages     = {2474--2483},
  year      = {2018},
  crossref  = {DBLP:conf/icml/2018},
  url       = {http://proceedings.mlr.press/v80/kang18a.html},
  bibsource = {dblp computer science bibliography, https://dblp.org}
}

@inproceedings{DBLP:journals/jmlr/RossGB11,
  author    = {St{\'{e}}phane Ross and
               Geoffrey J. Gordon and
               Drew Bagnell},
  title     = {A Reduction of Imitation Learning and Structured Prediction to No-Regret
               Online Learning},
  booktitle = {Proceedings of the Fourteenth International Conference on Artificial
               Intelligence and Statistics, {AISTATS} 2011, Fort Lauderdale, USA,
               April 11-13, 2011},
  pages     = {627--635},
  year      = {2011},
  crossref  = {DBLP:conf/aistats/2011},
  url       = {http://proceedings.mlr.press/v15/ross11a/ross11a.pdf},
  bibsource = {dblp computer science bibliography, https://dblp.org}
}

@article{DBLP:journals/corr/SallabSTA17,
  author    = {Ahmad El Sallab and
               Mahmoud Saeed and
               Omar Abdel Tawab and
               Mohammed Abdou},
  title     = {Meta learning Framework for Automated Driving},
  journal   = {CoRR},
  volume    = {abs/1706.04038},
  year      = {2017},
  url       = {http://arxiv.org/abs/1706.04038},
  archivePrefix = {arXiv},
  eprint    = {1706.04038},
  bibsource = {dblp computer science bibliography, https://dblp.org}
}

@inproceedings{DBLP:conf/icml/FinnAL17,
  author    = {Chelsea Finn and
               Pieter Abbeel and
               Sergey Levine},
  title     = {Model-Agnostic Meta-Learning for Fast Adaptation of Deep Networks},
  booktitle = {Proceedings of the 34th International Conference on Machine Learning,
               {ICML} 2017, Sydney, NSW, Australia, 6-11 August 2017},
  pages     = {1126--1135},
  year      = {2017},
  crossref  = {DBLP:conf/icml/2017},
  url       = {http://proceedings.mlr.press/v70/finn17a.html},
  bibsource = {dblp computer science bibliography, https://dblp.org}
}

@article{DBLP:journals/corr/abs-1801-06503,
  author    = {Alexandre Attia and
               Sharone Dayan},
  title     = {Global overview of Imitation Learning},
  journal   = {CoRR},
  volume    = {abs/1801.06503},
  year      = {2018},
  url       = {http://arxiv.org/abs/1801.06503},
  archivePrefix = {arXiv},
  eprint    = {1801.06503},
  bibsource = {dblp computer science bibliography, https://dblp.org}
}

@article{DBLP:journals/corr/abs-1812-00971,
  author    = {Mitchell Wortsman and
               Kiana Ehsani and
               Mohammad Rastegari and
               Ali Farhadi and
               Roozbeh Mottaghi},
  title     = {Learning to Learn How to Learn: Self-Adaptive Visual Navigation Using
               Meta-Learning},
  journal   = {CoRR},
  volume    = {abs/1812.00971},
  year      = {2018},
  url       = {http://arxiv.org/abs/1812.00971},
  archivePrefix = {arXiv},
  eprint    = {1812.00971},
  bibsource = {dblp computer science bibliography, https://dblp.org}
}

@inproceedings{DBLP:conf/globecom/XuLGKAW18,
  author    = {Junhong Xu and
               Qiwei Liu and
               Hanging Guo and
               Aaron Kageza and
               Saeed AlQarni and
               Shaoen Wu},
  title     = {Shared Multi-Task Imitation Learning for Indoor Self-Navigation},
  booktitle = {{IEEE} Global Communications Conference, {GLOBECOM} 2018, Abu Dhabi,
               United Arab Emirates, December 9-13, 2018},
  pages     = {1--7},
  year      = {2018},
  crossref  = {DBLP:conf/globecom/2018},
  url       = {https://doi.org/10.1109/GLOCOM.2018.8647614},
  doi       = {10.1109/GLOCOM.2018.8647614},
  bibsource = {dblp computer science bibliography, https://dblp.org}
}

@inproceedings{DBLP:conf/icml/0001JADYD18,
  author    = {Hoang Minh Le and
               Nan Jiang and
               Alekh Agarwal and
               Miroslav Dud{\'{\i}}k and
               Yisong Yue and
               Hal Daum{\'{e}} III},
  title     = {Hierarchical Imitation and Reinforcement Learning},
  booktitle = {Proceedings of the 35th International Conference on Machine Learning,
               {ICML} 2018, Stockholmsm{\"{a}}ssan, Stockholm, Sweden, July
               10-15, 2018},
  pages     = {2923--2932},
  year      = {2018},
  crossref  = {DBLP:conf/icml/2018},
  url       = {http://proceedings.mlr.press/v80/le18a.html},
  bibsource = {dblp computer science bibliography, https://dblp.org}
}

@article{DBLP:journals/ftrob/OsaPNBA018,
  author    = {Takayuki Osa and
               Joni Pajarinen and
               Gerhard Neumann and
               J. Andrew Bagnell and
               Pieter Abbeel and
               Jan Peters},
  title     = {An Algorithmic Perspective on Imitation Learning},
  journal   = {Foundations and Trends in Robotics},
  volume    = {7},
  number    = {1-2},
  pages     = {1--179},
  year      = {2018},
  url       = {https://doi.org/10.1561/2300000053},
  doi       = {10.1561/2300000053},
  bibsource = {dblp computer science bibliography, https://dblp.org}
}

@inproceedings{DBLP:conf/icra/EnglertPPD13,
  author    = {Peter Englert and
               Alexandros Paraschos and
               Jan Peters and
               Marc Peter Deisenroth},
  title     = {Model-based imitation learning by probabilistic trajectory matching},
  booktitle = {2013 {IEEE} International Conference on Robotics and Automation, Karlsruhe,
               Germany, May 6-10, 2013},
  pages     = {1922--1927},
  year      = {2013},
  crossref  = {DBLP:conf/icra/2013},
  url       = {https://doi.org/10.1109/ICRA.2013.6630832},
  doi       = {10.1109/ICRA.2013.6630832},
  bibsource = {dblp computer science bibliography, https://dblp.org}
}

@book{Quionero-Candela:2009:DSM:1462129,
 author = {Quionero-Candela, Joaquin and Sugiyama, Masashi and Schwaighofer, Anton and Lawrence, Neil D.},
 title = {Dataset Shift in Machine Learning},
 year = {2009},
 isbn = {0262170051, 9780262170055},
 publisher = {The MIT Press},
}

@inproceedings{DBLP:conf/corl/LaskeyLFDG17,
  author    = {Michael Laskey and
               Jonathan Lee and
               Roy Fox and
               Anca D. Dragan and
               Ken Goldberg},
  title     = {{DART:} Noise Injection for Robust Imitation Learning},
  booktitle = {1st Annual Conference on Robot Learning, CoRL 2017, Mountain View,
               California, USA, November 13-15, 2017, Proceedings},
  pages     = {143--156},
  year      = {2017},
  crossref  = {DBLP:conf/corl/2017},
  url       = {http://proceedings.mlr.press/v78/laskey17a.html},
  bibsource = {dblp computer science bibliography, https://dblp.org}
}

@inproceedings{DBLP:conf/uai/FoxPT16,
  author    = {Roy Fox and
               Ari Pakman and
               Naftali Tishby},
  title     = {Taming the Noise in Reinforcement Learning via Soft Updates},
  booktitle = {Proceedings of the Thirty-Second Conference on Uncertainty in Artificial
               Intelligence, {UAI} 2016, June 25-29, 2016, New York City, NY, {USA}},
  year      = {2016},
  crossref  = {DBLP:conf/uai/2016},
  url       = {http://auai.org/uai2016/proceedings/papers/219.pdf},
  bibsource = {dblp computer science bibliography, https://dblp.org}
}

@inproceedings{DBLP:conf/icml/RenZYU18,
  author    = {Mengye Ren and
               Wenyuan Zeng and
               Bin Yang and
               Raquel Urtasun},
  title     = {Learning to Reweight Examples for Robust Deep Learning},
  booktitle = {Proceedings of the 35th International Conference on Machine Learning,
               {ICML} 2018, Stockholmsm{\"{a}}ssan, Stockholm, Sweden, July
               10-15, 2018},
  pages     = {4331--4340},
  year      = {2018},
  crossref  = {DBLP:conf/icml/2018},
  url       = {http://proceedings.mlr.press/v80/ren18a.html},
  bibsource = {dblp computer science bibliography, https://dblp.org}
}

@inproceedings{DBLP:conf/iclr/GaoXLYLD18,
  author    = {Yang Gao and
               Huazhe Xu and
               Ji Lin and
               Fisher Yu and
               Sergey Levine and
               Trevor Darrell},
  title     = {Reinforcement Learning from Imperfect Demonstrations},
  booktitle = {6th International Conference on Learning Representations, {ICLR} 2018,
               Vancouver, BC, Canada, April 30 - May 3, 2018, Workshop Track Proceedings},
  year      = {2018},
  crossref  = {DBLP:conf/iclr/2018w},
  url       = {https://openreview.net/forum?id=HytbCQG8z},
  bibsource = {dblp computer science bibliography, https://dblp.org}
}

@article{DBLP:journals/corr/abs-1810-11043,
  author    = {Tianhe Yu and
               Pieter Abbeel and
               Sergey Levine and
               Chelsea Finn},
  title     = {One-Shot Hierarchical Imitation Learning of Compound Visuomotor Tasks},
  journal   = {CoRR},
  volume    = {abs/1810.11043},
  year      = {2018},
  url       = {http://arxiv.org/abs/1810.11043},
  archivePrefix = {arXiv},
  eprint    = {1810.11043},
  bibsource = {dblp computer science bibliography, https://dblp.org}
}

@inproceedings{DBLP:conf/rss/PanCSLYTB18,
  author    = {Yunpeng Pan and
               Ching{-}An Cheng and
               Kamil Saigol and
               Keuntaek Lee and
               Xinyan Yan and
               Evangelos A. Theodorou and
               Byron Boots},
  title     = {Agile Autonomous Driving using End-to-End Deep Imitation Learning},
  booktitle = {Robotics: Science and Systems XIV, Carnegie Mellon University, Pittsburgh,
               Pennsylvania, USA, June 26-30, 2018},
  year      = {2018},
  crossref  = {DBLP:conf/rss/2018},
  url       = {http://www.roboticsproceedings.org/rss14/p56.html},
  doi       = {10.15607/RSS.2018.XIV.056},
  bibsource = {dblp computer science bibliography, https://dblp.org}
}

@inproceedings{DBLP:conf/nips/HaS18,
  author    = {David Ha and
               J{\"{u}}rgen Schmidhuber},
  title     = {Recurrent World Models Facilitate Policy Evolution},
  booktitle = {Advances in Neural Information Processing Systems 31: Annual Conference
               on Neural Information Processing Systems 2018, NeurIPS 2018, 3-8 December
               2018, Montr{\'{e}}al, Canada.},
  pages     = {2455--2467},
  year      = {2018},
  crossref  = {DBLP:conf/nips/2018},
  url       = {http://papers.nips.cc/paper/7512-recurrent-world-models-facilitate-policy-evolution},
  bibsource = {dblp computer science bibliography, https://dblp.org}
}

@inproceedings{DBLP:conf/icra/CodevillaMLKD18,
  author    = {Felipe Codevilla and
               Matthias Miiller and
               Antonio L{\'{o}}pez and
               Vladlen Koltun and
               Alexey Dosovitskiy},
  title     = {End-to-End Driving Via Conditional Imitation Learning},
  booktitle = {2018 {IEEE} International Conference on Robotics and Automation, {ICRA}
               2018, Brisbane, Australia, May 21-25, 2018},
  pages     = {1--9},
  year      = {2018},
  crossref  = {DBLP:conf/icra/2018},
  url       = {https://doi.org/10.1109/ICRA.2018.8460487},
  doi       = {10.1109/ICRA.2018.8460487},
  bibsource = {dblp computer science bibliography, https://dblp.org}
}

@article{DBLP:journals/tamd/Schmidhuber10,
  author    = {J{\"{u}}rgen Schmidhuber},
  title     = {Formal Theory of Creativity, Fun, and Intrinsic Motivation {(1990-2010)}},
  journal   = {{IEEE} Trans. Autonomous Mental Development},
  volume    = {2},
  number    = {3},
  pages     = {230--247},
  year      = {2010},
  url       = {https://doi.org/10.1109/TAMD.2010.2056368},
  doi       = {10.1109/TAMD.2010.2056368},
  bibsource = {dblp computer science bibliography, https://dblp.org}
}

@Inbook{Schmidhuber1990,
author="Schmidhuber, J{\"u}rgen",
title="Reinforcement Learning with Interacting Continually Running Fully Recurrent Networks",
bookTitle="International Neural Network Conference: July 9--13, 1990 Palais Des Congres --- Paris --- France",
year="1990",
publisher="Springer Netherlands",
address="Dordrecht",
pages="817--820",
isbn="978-94-009-0643-3",
doi="10.1007/978-94-009-0643-3_97",
url="https://doi.org/10.1007/978-94-009-0643-3_97"
}

@article{DBLP:journals/corr/BojarskiTDFFGJM16,
  author    = {Mariusz Bojarski and
               Davide Del Testa and
               Daniel Dworakowski and
               Bernhard Firner and
               Beat Flepp and
               Prasoon Goyal and
               Lawrence D. Jackel and
               Mathew Monfort and
               Urs Muller and
               Jiakai Zhang and
               Xin Zhang and
               Jake Zhao and
               Karol Zieba},
  title     = {End to End Learning for Self-Driving Cars},
  journal   = {CoRR},
  volume    = {abs/1604.07316},
  year      = {2016},
  url       = {http://arxiv.org/abs/1604.07316},
  archivePrefix = {arXiv},
  eprint    = {1604.07316},
  bibsource = {dblp computer science bibliography, https://dblp.org}
}

@inproceedings{DBLP:conf/corl/FinnYZAL17,
  author    = {Chelsea Finn and
               Tianhe Yu and
               Tianhao Zhang and
               Pieter Abbeel and
               Sergey Levine},
  title     = {One-Shot Visual Imitation Learning via Meta-Learning},
  booktitle = {1st Annual Conference on Robot Learning, CoRL 2017, Mountain View,
               California, USA, November 13-15, 2017, Proceedings},
  pages     = {357--368},
  year      = {2017},
  crossref  = {DBLP:conf/corl/2017},
  url       = {http://proceedings.mlr.press/v78/finn17a.html},
  bibsource = {dblp computer science bibliography, https://dblp.org}
}

@inproceedings{DBLP:conf/nips/DuanASHSSAZ17,
  author    = {Yan Duan and
               Marcin Andrychowicz and
               Bradly C. Stadie and
               Jonathan Ho and
               Jonas Schneider and
               Ilya Sutskever and
               Pieter Abbeel and
               Wojciech Zaremba},
  title     = {One-Shot Imitation Learning},
  booktitle = {Advances in Neural Information Processing Systems 30: Annual Conference
               on Neural Information Processing Systems 2017, 4-9 December 2017,
               Long Beach, CA, {USA}},
  pages     = {1087--1098},
  year      = {2017},
  crossref  = {DBLP:conf/nips/2017},
  url       = {http://papers.nips.cc/paper/6709-one-shot-imitation-learning},
  bibsource = {dblp computer science bibliography, https://dblp.org}
}

@inproceedings{DBLP:conf/iclr/0004DLAL17,
  author    = {Abhishek Gupta and
               Coline Devin and
               Yuxuan Liu and
               Pieter Abbeel and
               Sergey Levine},
  title     = {Learning Invariant Feature Spaces to Transfer Skills with Reinforcement
               Learning},
  booktitle = {5th International Conference on Learning Representations, {ICLR} 2017,
               Toulon, France, April 24-26, 2017, Conference Track Proceedings},
  year      = {2017},
  crossref  = {DBLP:conf/iclr/2017},
  url       = {https://openreview.net/forum?id=Hyq4yhile},
  bibsource = {dblp computer science bibliography, https://dblp.org}
}

@inproceedings{DBLP:conf/icra/LekkalaI21,
  author       = {Kiran Lekkala and
                  Laurent Itti},
  title        = {Shaped Policy Search for Evolutionary Strategies using Waypoints\({}^{\mbox{*}}\)},
  booktitle    = {{IEEE} International Conference on Robotics and Automation, {ICRA}
                  2021, Xi'an, China, May 30 - June 5, 2021},
  pages        = {9093--9100},
  publisher    = {{IEEE}},
  year         = {2021},
  url          = {https://doi.org/10.1109/ICRA48506.2021.9561607},
  doi          = {10.1109/ICRA48506.2021.9561607},
  bibsource    = {dblp computer science bibliography, https://dblp.org}
}

@article{lekkala2020attentive,
  title={Attentive Feature Reuse for Multi Task Meta learning},
  author={Lekkala, Kiran and Itti, Laurent},
  journal={arXiv preprint arXiv:2006.07438},
  year={2020}
}

@article{DBLP:journals/corr/abs-2305-15591,
  author       = {Yunhao Ge and
                  Yuecheng Li and
                  Di Wu and
                  Ao Xu and
                  Adam M. Jones and
                  Amanda Sofie Rios and
                  Iordanis Fostiropoulos and
                  Shixian Wen and
                  Po{-}Hsuan Huang and
                  Zachary William Murdock and
                  Gozde Sahin and
                  Shuo Ni and
                  Kiran Lekkala and
                  Sumedh Anand Sontakke and
                  Laurent Itti},
  title        = {Lightweight Learner for Shared Knowledge Lifelong Learning},
  journal      = {CoRR},
  volume       = {abs/2305.15591},
  year         = {2023},
  url          = {https://doi.org/10.48550/arXiv.2305.15591},
  doi          = {10.48550/arXiv.2305.15591},
  eprinttype    = {arXiv},
  eprint       = {2305.15591},
  bibsource    = {dblp computer science bibliography, https://dblp.org}
}

@article{xu2022ferroelectric,
  title={Ferroelectric fet based context-switching fpga enabling dynamic reconfiguration for adaptive deep learning machines},
  author={Xu, Yixin and Zhao, Zijian and Xiao, Yi and Yu, Tongguang and Mulaosmanovic, Halid and Kleimaier, Dominik and Duenkel, Stefan and Beyer, Sven and Gong, Xiao and Joshi, Rajiv and others},
  journal={arXiv preprint arXiv:2212.00089},
  year={2022}
}

@inproceedings{lekkala2016accurate,
  title={Accurate and augmented navigation for quadcopter based on multi-sensor fusion},
  author={Lekkala, Kiran Kumar and Mittal, Vinay Kumar},
  booktitle={2016 IEEE Annual India Conference (INDICON)},
  pages={1--6},
  year={2016},
  organization={IEEE}
}

@inproceedings{lekkala2015artificial,
  title={Artificial intelligence for precision movement robot},
  author={Lekkala, Kiran Kumar and Mittal, Vinay Kumar},
  booktitle={2015 2nd International Conference on Signal Processing and Integrated Networks (SPIN)},
  pages={378--383},
  year={2015},
  organization={IEEE}
}

@inproceedings{lekkala2014pid,
  title={PID controlled 2D precision robot},
  author={Lekkala, Kiran Kumar and Mittal, Vinay Kumar},
  booktitle={2014 International Conference on Control, Instrumentation, Communication and Computational Technologies (ICCICCT)},
  pages={1141--1145},
  year={2014},
  organization={IEEE}
}

@inproceedings{lekkala2016simultaneous,
  title={Simultaneous aerial vehicle localization and human tracking},
  author={Lekkala, Kiran Kumar and Mittal, Vinay Kumar},
  booktitle={2016 IEEE Region 10 Conference (TENCON)},
  pages={379--383},
  year={2016},
  organization={IEEE}
}

@article{wen2022can,
  title={What can we learn from misclassified ImageNet images?},
  author={Wen, Shixian and Rios, Amanda Sofie and Lekkala, Kiran and Itti, Laurent},
  journal={arXiv preprint arXiv:2201.08098},
  year={2022}
}

\end{document}